\documentclass[letterpaper, 10 pt, conference]{ieeeconf}  % Comment this line out
\IEEEoverridecommandlockouts                              % This command is only
\usepackage{xurl}
\usepackage[utf8]{inputenc}
\usepackage[T1]{fontenc}
\usepackage{graphicx}
\usepackage{booktabs}
\usepackage{threeparttable}
\usepackage{graphics} % for pdf, bitmapped graphics files
\usepackage{epsfig} % for postscript graphics files
\usepackage{mathptmx} % assumes new font selection scheme installed
\usepackage{mathptmx} % assumes new font selection scheme installed
\usepackage{amsmath} % assumes amsmath package installed
\usepackage{amssymb}  % assumes amsmath package installed
\usepackage{xcolor}
\usepackage{multirow}
\usepackage{multicol}

\title{\LARGE \bf
Beyond 9-to-5: A Generative Model for Augmenting Mobility Data of Underrepresented Shift Workers
}

\author{Haoxuan Ma$^{1}$, Xishun Liao$^{1*}$, Yifan Liu$^{1}$, Chris Stanford$^{2}$, and Jiaqi Ma$^{1}$% <-this % stops a space
\thanks{$^{1}$Department of Civil and Environmental Engineering, Samueli School of Engineering, University of California at Los Angeles, Los Angeles, CA 90095}% <-this % stops a space
\thanks{$^{2}$Novateur Research Solutions, 20110 Ashbrook Place, STE 170, Ashburn, 20147, VA, USA}%
\thanks{$^{*}$Corresponding Author: Xishun Liao (e-mail: xishunliao@ucla.edu)}%
}

\begin{document}

\maketitle
\thispagestyle{empty}
\pagestyle{empty}

\begin{abstract}
% This paper addresses a critical gap in urban mobility modeling by focusing on shift workers, a population segment comprising 15-20\% of the workforce yet systematically underrepresented in traditional transportation surveys. Comparative analysis of GPS and survey data reveals stark differences between shift workers' bimodal temporal patterns and conventional 9-to-5 schedules. We introduce a novel transformer-based approach leveraging fragmented GPS trajectory data to generate complete, behaviorally valid activity patterns for individuals working non-standard hours. Our method employs period-aware temporal embeddings and a transition-focused loss function to capture unique shift worker activity rhythms and mitigate biases in conventional datasets. Evaluation demonstrates remarkable distributional alignment with Los Angeles County GPS data (Average JSD $<$ 0.02 for all metrics). By transforming incomplete GPS traces into complete activity patterns, our approach provides transportation planners with a data augmentation tool for understanding 24/7 urban mobility needs, enabling more inclusive transportation planning.
This paper addresses a critical gap in urban mobility modeling by focusing on shift workers, a population segment comprising 15-20\% of the workforce in industrialized societies yet systematically underrepresented in traditional transportation surveys and planning. This underrepresentation is revealed in this study by a comparative analysis of GPS and survey data, highlighting stark differences between the bimodal temporal patterns of shift workers and the conventional 9-to-5 schedules recorded in surveys. To address this bias, we introduce a novel transformer-based approach that leverages fragmented GPS trajectory data to generate complete, behaviorally valid activity patterns for individuals working non-standard hours. Our method employs periodaware temporal embeddings and a transition-focused loss function specifically designed to capture the unique activity rhythms of shift workers and mitigate the inherent biases in conventional transportation datasets. Evaluation shows that the generated data achieves remarkable distributional alignment with GPS data from Los Angeles County (Average JSD < 0.02 for all evaluation metrics). By transforming incomplete GPS traces into complete, representative activity patterns, our approach provides transportation planners with a powerful data augmentation tool to fill critical gaps in understanding the 24/7 mobility needs of urban populations, enabling precise and inclusive transportation planning.
\end{abstract}

\section{INTRODUCTION}

Human mobility modeling is critical for urban planning, transportation management, and public policy development \cite{gonzalez2008understanding, liao2024deep, stanford2024numosim}. Accurate representation of how individuals interact with urban environments enables planners to design efficient, equitable, and sustainable cities \cite{hasanSpatiotemporal2013, maAI2024, ma2024ev, martin2025survey}. However, shift workers operating outside traditional daytime hours remain understudied despite comprising a significant urban population segment.

Shift workers in healthcare, manufacturing, emergency services, and service industries constitute approximately 15–20\% of the workforce \cite{eurofound2016ewcs, NBERw20449, apta2019lateshift}. Their mobility patterns differ markedly from daytime workers, characterized by activities crossing midnight boundaries and reduced discretionary travel \cite{palmImpact2024}. Limited overnight transportation services create unique accessibility challenges, disproportionately affecting lower-income and minority shift workers \cite{apta2019lateshift, palmShifted2023, tpb2019latenight}.

Shift worker underrepresentation stems from two interconnected issues. First, conventional household surveys undersample non-standard schedule workers, struggling to capture overnight mobility behaviors \cite{inbook, lim2024shift}. Second, traditional activity-based models rely on these biased datasets and focus on within-day patterns, overlooking midnight-crossing activities that characterize shift work \cite{bastarianto2023agent, bhatActivity1999, bowmanABM2001}.

GPS-enabled devices offer opportunities to supplement biased survey data through continuous, high-resolution mobility traces. However, GPS datasets present challenges including temporal inconsistency, coverage fluctuations, and varying reliability \cite{SilaNowicka2016, Palmer2013}.

We propose a transformer-based framework specifically for shift workers. The self-attention mechanism learns temporal relationships between observed activities to infer missing ones, converting partial GPS traces into complete activity diaries. This enables accurate capture of unique shift worker temporal patterns while serving as a data augmentation tool for realistic mobility trajectory generation. Applications include travel demand modeling, transit scheduling, signal control, and MaaS platform optimization for equitable shift worker mobility \cite{rocco2024Transit,li2025Signal, othman2025signal}. Our contributions are:

\begin{itemize}
\item A transformer-based model that reconstructs GPS data into complete shift worker activity chains, mitigating survey biases and creating augmented datasets that rebalance systemic undersampling in plannings.

\item Extensive empirical evaluation using Los Angeles County GPS data, demonstrating synthetic yet behaviorally valid activity chain generation reflecting unique shift worker mobility dynamics.

\item Comparative analysis with traditional datasets and daytime worker patterns, highlighting shift workers' distinctive mobility needs and implications for inclusive transportation planning.
\end{itemize}

Our approach provides transportation planners with essential tools for understanding and accommodating shift worker mobility needs.

\section{Literature Review}

\subsection{Human Mobility Modeling}
Early transportation forecasting relied on trip-based models, notably the four-step travel demand model. These models assume average travel patterns and struggle with individual heterogeneity \cite{bastarianto2023agent}. Activity-based and agent-based models (ABM) were developed to simulate disaggregate travel \cite{bhatActivity1999, benakivaHybrid2002}. Bowman and Ben-Akiva established ABM frameworks using discrete choice models \cite{bowmanABM2001}, forming foundations for large-scale traffic simulation \cite{heABMTRANS}. However, traditional ABMs require extensive calibration and focus on within-day patterns, neglecting overnight transitions characterizing shift worker mobility.

Ubiquitous data sources—GPS devices, smartphones, transit cards, and sensors—have enabled data-driven mobility modeling where machine learning infers travel behavior from large-scale traces \cite{zhao2018Smartcard}. Deep learning techniques including RNN, LSTM, GAN, and Diffusion Models have achieved prominence in mobility modeling \cite{luca2021surveydeeplearninghuman}, tackling problems from next-trip prediction to trajectory generation and often outperforming traditional models in capturing spatiotemporal patterns. Generative models now synthesize realistic mobility patterns, offering data-driven alternatives to rule-based simulators \cite{liao2024deep, ma2025mobifuse, giulio2024synth}.

\subsection{Shift Worker Mobility Research}
Research examining shift worker mobility remains limited, focusing primarily on documenting challenges rather than developing modeling solutions. Palm \cite{palmImpact2024} quantified how night shift workers make significantly fewer discretionary trips than daytime workers. This aligns with research showing non-standard schedules create misalignment with social rhythms and reduce community participation \cite{palmShifted2023}. Existing studies rely on traditional surveys with sampling biases \cite{inbook, lim2024shift}, qualitative interviews, or aggregate transit data obscuring individual patterns \cite{tpb2019latenight}. Lim et al. \cite{lim2024shift} identified distinct night-shift commuting patterns (fixed vs. rotating), highlighting modeling complexity. Kapitza \cite{KAPITZA2022103418} demonstrated nighttime increases car dependency with gender and urbanity effects, while Jang \cite{jang2023working} showed shift workers work longer hours than day workers. Despite these insights, comprehensive temporal modeling approaches for shift workers' unique spatiotemporal patterns remain lacking. Transportation barriers underscore needs for specialized approaches \cite{tpb2019latenight, apta2019lateshift}.

\subsection{Research Gaps and Our Approach}
Three key gaps exist in shift worker mobility modeling: (1) current models fail to capture midnight-crossing activities, (2) temporal inconsistency and missing GPS data particularly affect overnight periods, and (3) heterogeneity among shift worker types requires preservation. Our transformer-based approach addresses these gaps by analyzing fragmented GPS trajectories to generate complete, behaviorally valid activity patterns capturing unique shift worker temporal characteristics. Agent-level processing preserves distinctive shift work schedule features while reconstructing complete activity chains from incomplete observations, enabling better understanding of this underrepresented workforce segment.

\section{Problem Formulation}

We focus on developing a data augmentation approach for overnight and shift worker mobility patterns to address the limitations of current transportation models and data collection methods. Our approach aims to generate complete, realistic activity sequences that accurately capture the unique temporal characteristics of shift workers whose activities span evening hours and cross midnight boundaries.

We denote $i$ for an agent (individual). For each agent, we collect a pair of consecutive daily activity chains spanning two days. The activity chain for day $d$ (where $d \in \{1, 2\}$) is represented as $A_d^i = \{a_1^{i,d}, a_2^{i,d}, \ldots, a_{N_d}^{i,d}\}$, where $N_d$ is the number of activities performed on day $d$.

Each activity $a_n^{i,d} = [T_n^{i,d}, S_n^{i,d}, E_n^{i,d}]$ consists of:
Activity type $T_n^{i,d}$ (from the set of 15 activity categories in Table \ref{table:activity_type_table}), Start time $S_n^{i,d}$ (represented as time slot within the day), End time $E_n^{i,d}$ (represented as time slot within the day).
For computational efficiency, we discretize each 24-hour day into 96 time slots, each representing a 15-minute interval. Therefore, $S_n^{i,d}, E_n^{i,d} \in \{1,2,...,96\}$.

Given an observed activity pattern, our objective is to develop a model $f_{\theta}$ that generates activity with precise temporal characteristics of shift workers, including accurate work-sleep cycles, appropriate activity durations, and properly captured midnight-crossing activities: $\hat{A} = f_{\theta}(A_{obs})$.

To address the critical challenge of fragmented GPS data, we implement a masking approach that explicitly models these observation gaps. We denote the observation mask for day $d$ as $M_d^i$, where each time slot $t$ has a corresponding mask value $M_d^i[t] \in \{0,1\}$, with 1 indicating an observed time slot and 0 indicating a gap in observation. This masking mechanism is a key of our solution, enabling our model to learn from incomplete data and generate complete activity chains despite the inherent limitations of GPS trajectories.

Given a dataset $\mathcal{D} = \{(A_1^i, A_2^i, M_1^i, M_2^i) | i \in \{1,...,I\}\}$ of paired activity chains from $I$ agents, we aim to train a model that minimizes the discrepancy between generated and actual activities while accounting for observation gaps: $\min_{\theta} \sum_{i=1}^{I} \mathcal{L}(f_{\theta}(A_1^i), A_2^i, M_2^i)$, where $\mathcal{L}$ is a suitable loss function that evaluates generation accuracy only on observed time slots, places special emphasis on correctly predicting activity transitions, and preserves the temporal patterns characteristic of shift workers.

The trained model serves as a data augmentation tool to bridge gaps in conventional survey data and address the temporal inconsistency issues in GPS trajectories, ultimately enabling more accurate representation of shift worker patterns for transportation planning and policy analysis.

{\setlength{\abovecaptionskip}{0pt}%
\setlength{\belowcaptionskip}{0pt}%
\setlength{\textfloatsep}{0pt}%
\begin{table}[ht]
    \centering
    \small
    \caption{Activity category codes with descriptions \cite{nhtsdata}}
    \begin{tabular}{|l|l|l|}
        \hline
        1. Home           & 6. Shop services  & 11. Social      \\ \hline
        2. Work           & 7. Meals out      & 12. Healthcare  \\ \hline
        3. School         & 8. Errands        & 13. Worship     \\ \hline
        4. Caregiving     & 9. Leisure        & 14. Other       \\ \hline
        5. Shop goods     & 10. Exercise      & 15. Pickup/Drop \\ \hline
    \end{tabular}
    \label{table:activity_type_table}
\end{table}}

\section{Methodology}

\subsection{System Workflow}

\begin{figure}[t]
    \centering
    \includegraphics[width=0.9\columnwidth]{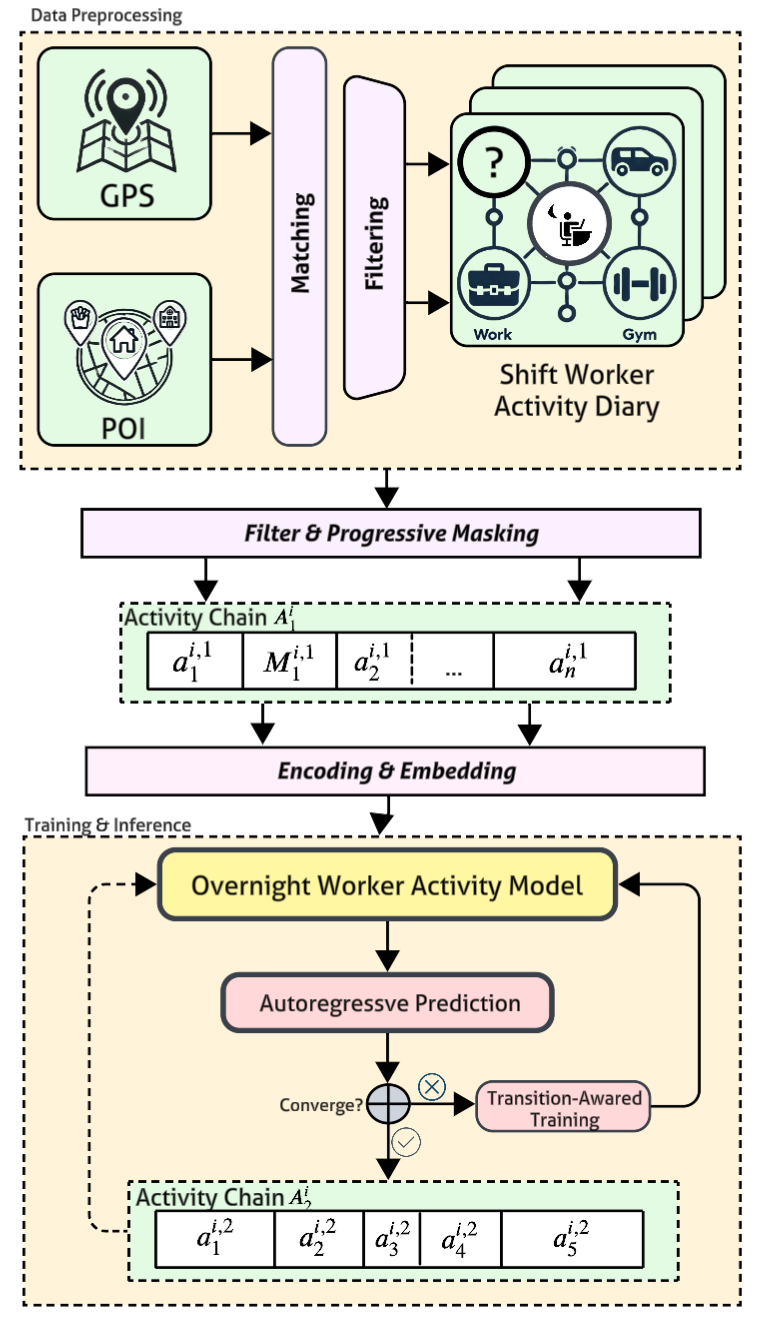}
    \caption{System Workflow of the Proposed Model}
    \label{fig:system_diagram}
\end{figure}

The shift worker mobility generation system addresses the unique challenges of modeling shift worker activity patterns, which have been traditionally underrepresented in transportation planning. Our workflow, illustrated in Fig. \ref{fig:system_diagram}, encompasses data processing, model training, and inference stages specifically designed for capturing overnight activity transitions. The process begins with the preparation of paired activity chains from consecutive days, with special attention to evening and overnight periods. We processed the GPS points and constructed corresponding masking tensors to address the inherent gaps in GPS data collection, ensuring the model learns from reliable observations while appropriately handling missing data.

Our training pipeline incorporates specialized data augmentation techniques that emphasize realistic overnight activity transitions. The model undergoes training with a progressive masking schedule, where the masking ratio increases gradually to enhance the model's ability to infer activities from increasingly sparse observations. This approach mimics real-world scenarios where GPS data collection often suffers from irregular gaps, particularly during overnight periods.

A key innovation in our workflow is the transition-aware training mechanism that places greater emphasis on activity transitions rather than static periods. This addresses the tendency of conventional models to predict overly simplified activity patterns that fail to capture the complex temporal structure of shift worker schedules. The model receives stronger learning signals when predicting transitions between activities, encouraging it to develop more nuanced representations of activity change points. During inference, the trained model generates next-day activity patterns autoregressively, with each prediction conditioned on previously generated activities and the encoded representation of the first day. This sequential generation approach ensures temporal coherence across the generated activity chain.

\subsection{Model Structure}

\begin{figure}[t]
    \centering
    \includegraphics[width=0.9\columnwidth]{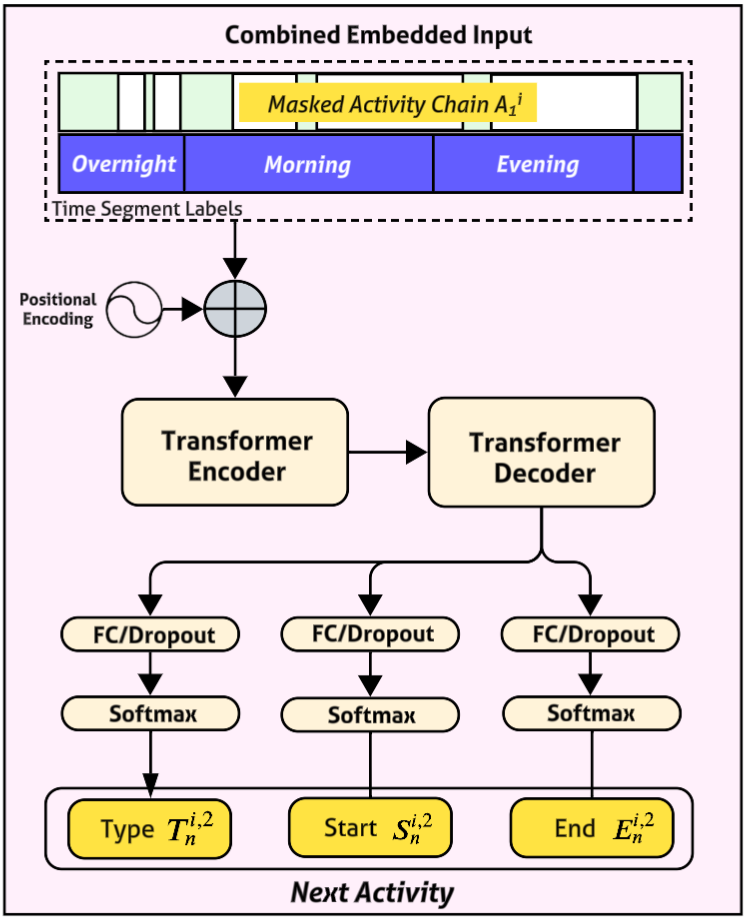}
    \caption{Network Architecture of the Model}
    \label{fig:network_architecture}
\end{figure}

We employ a transformer-based architecture optimized for activity generation, as illustrated in Fig. \ref{fig:network_architecture}.

\subsubsection{Combined Embedding Layer}
The embedding layer fuses activity type information with temporal context:
$\mathbf{e}_t = \mathbf{e}_{\text{act}}(x_t) + \mathbf{e}_{\text{time}}(t)$
where:
$\mathbf{e}_{\text{act}}(x_t) \in \mathbb{R}^d$ is the activity type embedding at time $t$, and
$\mathbf{e}_{\text{time}}(t) \in \mathbb{R}^d$ is the period-aware time embedding.

The period-aware time embedding is a key innovation that captures the temporal patterns of shift workers. It contains distinct time periods: $\mathbf{e}_{\text{time}}(t) = \mathbf{e}_{\text{pos}}(t) + \mathbf{e}_{\text{period}}(p(t)) + \mathbf{e}_{\text{sin}}(t)$
where: $\mathbf{e}_{\text{pos}}(t)$ is a standard positional embedding,  $\mathbf{e}_{\text{period}}(p(t))$ encodes the period as specified in below, and $\mathbf{e}_{\text{sin}}(t)$ adds sinusoidal features representing time of day.

The period function $p(t)$ is defined as:
$$p(t) =
\begin{cases}
\text{evening\_start}, & \text{if } t \in [18\cdot4, 22\cdot4) \\
\text{overnight}, & \text{if } t \in [22\cdot4, 6\cdot4) \\
\text{morning}, & \text{if } t \in [6\cdot4, 10\cdot4) \\
\text{other}, & \text{otherwise}
\end{cases}$$

For each sequence in the batch, positional indices are created and expanded to match the sequence dimensions. The activity embeddings and time embeddings are then combined through addition to form the final representation that captures both the activity type and its temporal context.

\subsubsection{Transformer Architecture}
Our model utilizes a transformer encoder-decoder architecture with modifications specifically tailored for activity sequence modeling. The encoder processes the embedded day 1 activities to create context-aware representations that capture temporal dependencies within the input sequence. The decoder then autoregressively generates day 2 activities one time slot at a time. To improve training stability and generation quality, we employ teacher forcing with probability $p_{tf}$, where ground truth values are occasionally used as decoder inputs instead of predicted values. The final output layer maps the decoder representations to activity type probabilities through a softmax function. Our implementation uses 4 layers in both encoder and decoder, with 8 attention heads and hidden dimension of 128, striking a balance between model capacity and computational efficiency.

\subsection{Loss Function Design}
Our loss function design addresses the unique challenges of shift worker activity generation through several complementary components.

\subsubsection{Cross-Entropy Loss}
The foundation of our loss function is the standard cross-entropy loss that measures the discrepancy between predicted and target activity types: $\mathcal{L}_{CE}(\hat{\mathbf{Y}}, \mathbf{Y}) = -\sum_{t=1}^{96} \log(\hat{\mathbf{Y}}_{t,y_t})$
where $y_t$ is the target class index at time $t$.

\subsubsection{Transition-Aware Loss}
To capture the temporal structure of activity patterns, we introduce a transition-aware loss that emphasizes correct prediction of activity transitions: $\mathcal{L}_{trans} = 1 - \mathrm{F1}_{transitions}$
This component uses precision and recall of transitions with a tolerance window to accommodate slight temporal shifts, rewarding the model for correctly predicting when activities change.

\subsubsection{Distribution Matching Loss}
To ensure the overall distribution of activities is realistic, we include a distribution matching component based on JSD (Jensen-Shannon divergence):
$\mathcal{L}_{dist} = JS(P_{\hat{\mathbf{Y}}} \parallel P_{\mathbf{Y}})$

JSD is a symmetric measure of similarity between two probability distributions, defined as:
$$JS(P \parallel Q) = \frac{1}{2}KL(P \parallel M) + \frac{1}{2}KL(Q \parallel M)$$
where $M = \frac{1}{2}(P + Q)$ is the midpoint distribution, and $KL(P \parallel Q) = \sum_i P(i) \log \frac{P(i)}{Q(i)}$ is the Kullback-Leibler divergence.

In our context, $P_{\hat{\mathbf{Y}}}$ represents the distribution of generated activities across all time slots, and $P_{\mathbf{Y}}$ represents the distribution of true activities. This encourages the model to maintain appropriate proportions of different activity types, preventing overrepresentation of common activities while ensuring rare activities are still captured in the generated sequences.

\subsubsection{Soft Label Loss}
The soft label component provides smoother supervision around transition points by using weighted combinations of activity types:
$\mathcal{L}_{soft} = -\sum_t \sum_c \tilde{y}_{t,c} \log(\hat{y}_{t,c})$
where $\tilde{y}_{t,c}$ are soft targets that blend between adjacent activities near transition points, allowing the model to learn more nuanced representations of activity changes.

Our final loss function combines these components with configurable weights:
$$\mathcal{L}_{combined} = \mathcal{L}_{CE} + \alpha \cdot \mathcal{L}_{trans} + \beta \cdot \mathcal{L}_{dist} + \gamma \cdot \mathcal{L}_{soft}$$
All components incorporate masking to handle observation gaps, ensuring the model is only trained on valid data points while preserving the temporal continuity of activity patterns.

\section{Experiment}

\subsection{Data Overview and Preprocess}
\paragraph{Comparative Analysis of Survey Data Under-representation} The examination of the activity distribution histograms in Table \ref{tab:activity_comparison} reveals significant discrepancies between the household travel survey (HTS) and GPS-derived activity patterns. While both datasets contain a mix of shift and 9-to-5 workers, the magnitude of early/late hour representation differs dramatically. For work activity start times, the HTS data shows substantially lower early morning work starts (0-6 AM) at only 7.3\% compared to 18.5\% in the GPS data. Similarly, evening work starts (6-12 PM) shows 4.2\% in HTS versus 7.5\% in GPS data. The home activity distributions further reinforce this pattern: morning home arrivals (6-12 AM), which would typically correspond to shift workers returning home, show a striking disparity with only 4.6\% captured in HTS compared to 15.4\% in GPS data. These systematic discrepancies across both work and home activities confirm our hypothesis that traditional transportation surveys significantly underrepresent non-standard schedules, particularly those of shift workers who comprise an essential segment of the urban workforce. The predominance of conventional 6-12 AM work starts (61.1\% in HTS vs. 52.2\% in GPS) and evening home returns (6-12 PM) in survey data demonstrates how traditional data collection methods skew toward capturing standard 9-to-5 work schedules while systematically missing shift worker mobility patterns.

\begin{table}[ht]
\centering
\caption{Comparison of activity start time distributions between traditional survey (HTS) and GPS data}
\label{tab:activity_comparison}
\begin{tabular}{|l|cc|cc|}
\hline
\multirow{2}{*}{\textbf{Time Period}} & \multicolumn{2}{c|}{\textbf{Work Activity (\%)}} & \multicolumn{2}{c|}{\textbf{Home Activity (\%)}} \\
\cline{2-5}
 & \textbf{HTS} & \textbf{GPS} & \textbf{HTS} & \textbf{GPS} \\
\hline
0-6 AM & 7.3 & 18.5 & 40.2 & 35.3 \\
6-12 AM & 61.1 & 52.2 & 4.6 & 15.4 \\
12-6 PM & 27.4 & 21.8 & 24.4 & 25.3 \\
6-12 PM & 4.2 & 7.5 & 30.7 & 24.0 \\
\hline
\end{tabular}
\end{table}

\paragraph{GPS Dataset} We utilize Los Angeles County GPS trajectory data collected over six months, preprocessed to extract activity patterns through stay point extraction and semantic enrichment via POI datasets annotated with activity types using large language models \cite{liuSemantic2024}. 

Shift workers are identified using three criteria: (1) work activities during evening hours (18:00-22:00), (2) work activities spanning midnight boundaries, and (3) sustained work periods during typical sleeping hours (22:00-06:00). This process identified 208,350 pairwise activity sequences representing two consecutive days from agents with shift work patterns.

\subsection{Experimental Setup}
Our transformer model was implemented in PyTorch and trained on an NVIDIA L40S GPU. We partitioned the dataset using an 80-10-10 split, resulting in 166,680 sequences for training, 20,835 for validation, and 20,835 for testing. The hyperparameters of the training are as following: training for 50 epochs with a batch size of 256, Adam optimizer with an initial learning rate of $1 \times 10^{-4}$, weight decay set to $1 \times 10^{-5}$ for regularization, gradient clipping with a threshold of 1.0, and dropout rate of 0.1 applied to all layers.

The core architecture consisted of a transformer encoder-decoder with 4 layers each, 8 attention heads, and a model dimension of 128. We utilized period-aware temporal embeddings specifically designed to capture the distinct patterns of shift workers, with particular attention to evening start (18:00-22:00), overnight (22:00-06:00), and morning transition (06:00-10:00) periods.
For our loss function, we implemented a transition-aware component with tolerance parameter $\tau=2$, allowing transitions to be predicted within a small window of time slots. This approach specifically addresses the challenge of capturing activity transitions accurately, rather than simply predicting static activities.

\subsection{Evaluation Methodology}
Evaluating human mobility models presents unique challenges compared to standard generation tasks. Individual-level accuracy metrics are often inadequate due to the inherent stochasticity and variability in human behavior. Instead, we adopt a distribution-based evaluation approach that assesses whether the model captures the statistical properties of human mobility at a population level.
We evaluate our model by comparing the distributions of various mobility characteristics between generated and real-world shift worker sequences. This approach allows us to assess whether the model has learned the underlying patterns governing shift worker mobility. For quantitative comparison, we employ JSD, as already expanded in the methodology section. JSD values range from 0 to 1, with 0 indicating identical distributions and values closer to 0 representing better matches.

We evaluate our model by comparing the distributions of various mobility characteristics between generated and real-world shift worker sequences. This approach allows us to assess whether the model has learned the underlying patterns governing shift worker mobility. For quantitative comparison, we employ JSD, as already expanded in the methodology section. To demonstrate the superiority of our model architecture, we also conducted benchmark comparisons against an LSTM with Attention model trained on identical data sources. We evaluate the following distributional characteristics: (1) temporal distributions of activity start and end times, (2) activity duration distributions, (3) activity type frequency distributions, and (4) work-specific temporal distributions. Additionally, we present a comparative analysis with 9-to-5 workers and NHTS (National Household Travel Survey) data to demonstrate the distinctive mobility patterns of shift workers and validate our model's ability to capture these unique characteristics.

\subsection{Results and Analysis} 
\subsubsection{Overall Distribution Alignment}

\begin{figure}[t]  % Use figure* to span two columns and place at top of page
    \centering
    \includegraphics[width=0.5\textwidth, keepaspectratio=false]{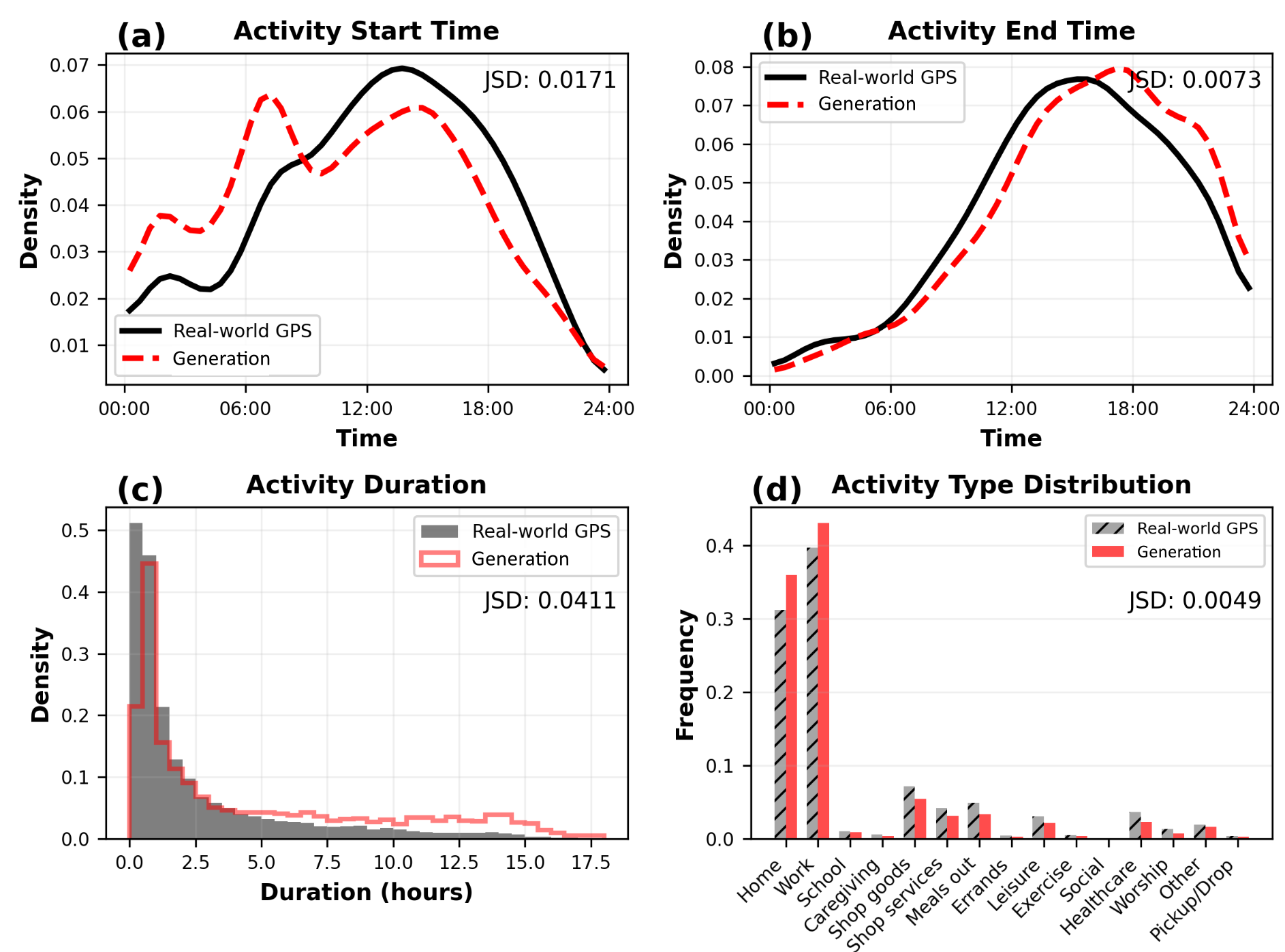}  % Control both width and height
    \caption{Comparison of distributional characteristics between generated and real-world mobility patterns}
    \label{fig:dist_diagram}
\end{figure}
Fig. \ref{fig:dist_diagram} presents the distributional comparison between our model's generation and real-world GPS for the test set. In Fig. \ref{fig:dist_diagram} (a) and (b), we observe a strong alignment in activity start time (JSD = 0.0171) and end time (JSD = 0.0073) distributions. This low divergence indicates that our model effectively captures the temporal dynamics of activity transitions throughout the day. The end time distribution shows particularly strong alignment, suggesting the model has learned realistic activity durations and transition patterns.

% Fig. \ref{fig:dist_diagram} (c) demonstrates good correspondence in activity duration distributions (JSD = 0.0411). We observe a slight tendency in the generated distribution to favor longer activities while underrepresenting very short activities. This phenomenon can be attributed to the inherent limitations of GPS data, where signal interruptions often fragment continuous activities into shorter segments with gaps. Conversely, the model learns to predict more realistic continuous activities, which better represents actual human behavior patterns rather than being influenced by data collection artifacts.

Fig. \ref{fig:dist_diagram} (c) demonstrates good correspondence in activity duration distributions (JSD = 0.0411). We observe a slight tendency in the generated distribution to favor longer activities while underrepresenting very short activities. This phenomenon can be attributed to the inherent limitations of GPS data, where signal interruptions often fragment continuous activities into shorter segments with gaps. Conversely, the model learns to predict more realistic continuous activities, which better represents actual human behavior patterns rather than being influenced by data collection artifacts. This finding aligns with Palm's research \cite{palmImpact2024}, which revealed that shift workers are less likely to engage in discretionary trips, typically have shorter duration than mandatory activities, compared to 9-to-5 workers.

The activity type distribution shown in Fig. \ref{fig:dist_diagram} (d) exhibits excellent alignment (JSD = 0.0049) between generated and real-world GPS frequencies. Mandatory activities show slightly higher prediction rates, while non-mandatory activities are marginally under-predicted. This asymmetry reflects the challenge in modeling discretionary activities, which exhibit greater variability and are influenced by contextual factors not captured in our input features.

Table~\ref{tab:jsd_comparison} presents the JSD metrics comparing our model against an LSTM with Attention baseline. The results demonstrate the exceptional performance of our approach across all evaluation dimensions. Our model achieves lower divergence from real-world GPS distributions, particularly in the activity start and end time distributions. Overall, our model's average JSD (0.0176) is significantly lower than the LSTM baseline (0.0621), confirming that the transformer architecture with period-aware embeddings and transition-focused loss is substantially more effective at capturing shift worker mobility patterns. The consistently lower JSD values across all metrics validate our architectural design choices and demonstrate our model's ability to generate highly realistic activity sequences that closely match real-world shift worker behaviors.

\begin{table}[h]
\centering
\caption{JSD Comparison Between Models Against GPS Data}
\begin{tabular}{lcc}
\hline
\textbf{Metric} & \textbf{Our Model} & \textbf{LSTM w/ Attn} \\
\hline
Start Time & \underline{0.0171} & 0.0653 \\
End Time & \underline{0.0073} & 0.0752 \\
Duration & \underline{0.0411} & 0.0800 \\
Activity Type & \underline{0.0049} & 0.0280 \\
\hline
\textbf{Average JSD} & \underline{0.0176} & 0.0621 \\
\hline
\end{tabular}
\label{tab:jsd_comparison}
\end{table}

\subsubsection{Work Activity Pattern Analysis}
\begin{figure}[t]
    \centering
    \includegraphics[width=\columnwidth]{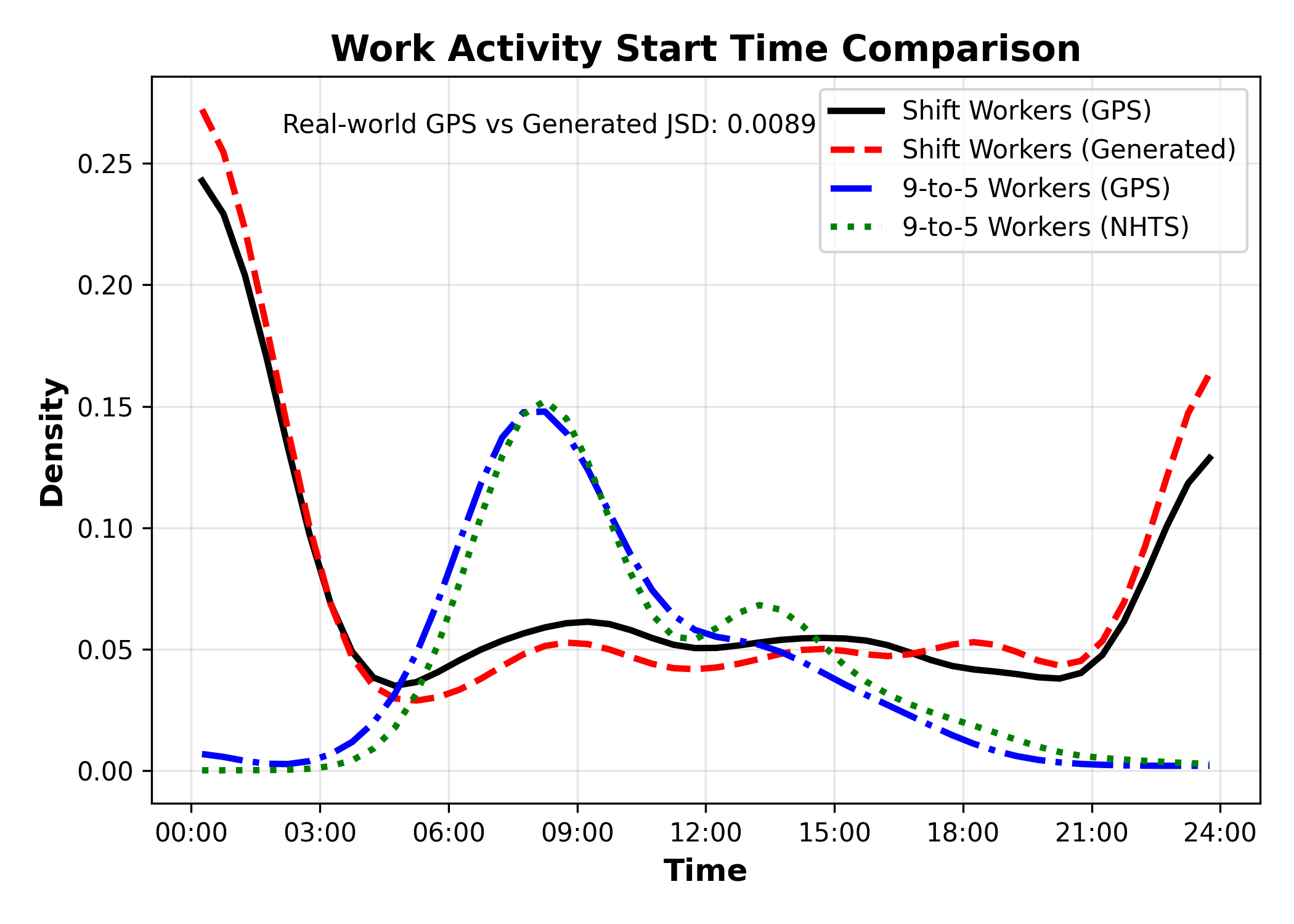}
    \caption{Comparison of work activity patterns between different worker types}
    \label{fig:worker_diagram}
\end{figure}

Fig. \ref{fig:worker_diagram} specifically analyzes the temporal distribution of work activities, which are particularly relevant for our target population. The work activity start time distribution (JSD = 0.0089) reveals distinctive bimodal peaks at early morning (00:00-03:00) and late evening (21:00-24:00) hours, characteristic of overnight shift schedules. Fig. \ref{fig:worker_diagram} also presents a comparative analysis of work activity patterns across different data sources: our generated shift worker patterns, real-world GPS overnight patterns, normal worker patterns from our dataset, and NHTS worker patterns. The stark contrast between overnight and regular worker patterns is immediately apparent. While normal workers and NHTS data show concentrated work start times in morning hours (07:00-09:00), shift workers exhibit peaks during late evening and early morning hours. The generated distributions (red dashed lines) closely track the real-world GPS distributions (black solid lines) for both work start and end times, capturing the characteristic overnight peaks at both ends of the day. This comparison validates our model's ability to reproduce the distinctive temporal signatures of shift workers accurately.

The clear divergence between shift worker patterns and those of normal workers (blue dash-dot lines) and NHTS workers (green dotted lines) in Fig. \ref{fig:worker_diagram} further demonstrates this bias. The latter two groups show nearly identical distributions with pronounced morning peaks for work start times and afternoon peaks for work end times, confirming that traditional surveys predominantly capture conventional 9-to-5 work schedules. This visualization powerfully demonstrates how transportation planning based solely on traditional survey data would systematically exclude the mobility needs of shift workers.

These results validate our approach to modeling shift worker mobility patterns and highlight the importance of developing specialized models for this significant yet often overlooked population. By supplementing traditional survey data with synthetically generated shift worker patterns, our model helps create more inclusive and representative datasets for transportation analysis, ensuring that the needs of all worker populations are considered in infrastructure and service planning The low JSD values across multiple distribution comparisons indicate that our transformer-based approach with period-aware embeddings and transition-focused loss functions effectively captures the complex temporal dynamics unique to shift workers.

\vspace{-1ex}

\section{Conclusion}

This paper addresses a critical gap in urban mobility modeling by developing a transformer-based approach specifically for shift workers. Our model achieves remarkable predictive accuracy, with JSD values below 0.02 for temporal distributions, validating its effectiveness in generating realistic activity chains for this underrepresented population.

The model effectively addresses fundamental challenges in GPS trajectory data through period-aware embeddings and masking mechanisms that infer activities during data gaps, creating complete and behaviorally valid activity chains. This capability transforms fragmentary mobility traces into comprehensive activity diaries suitable for transportation planning applications, enhancing representation of populations traditionally undersampled in conventional surveys.

\textit{Limitations:} Our GPS data inherently undersamples certain demographics due to smart device accessibility barriers. Additionally, activity detection relies on GPS-POI matching with rule-based assumptions \cite{liuSemantic2024}, introducing potential biases. The fragmented nature of GPS means we cannot identify all shift workers, particularly those with irregular schedules. However, our model's adaptability enables leveraging improved datasets without architectural changes, and our framework's reliance on widely available GPS and POI data enables deployment across diverse urban environments.

Beyond technical merits, this work serves as a call to action for transportation planning communities to recognize shift workers—an essential workforce keeping cities functioning around the clock yet systematically overlooked in policy decisions. Our approach provides planners with tools for understanding 24/7 urban mobility needs, enabling more inclusive transportation planning. Future work could model seasonal variations, incorporate weather context, and integrate with spatial models for comprehensive transportation demand forecasting, ultimately contributing to more equitable urban mobility systems serving all residents regardless of work schedules.

\section{Acknowledgement}
This work was supported in part by the FHWA Center for Excellence on New Mobility and Automated Vehicles Program, and in part by the Intelligence Advanced Research Projects Activity (IARPA) via Department of Interior/Interior Business Center (DOI/IBC) contract number 140D0423C0033. The U.S. Government is authorized to reproduce and distribute reprints for Governmental purposes notwithstanding any copyright annotation thereon. Disclaimer: The views and conclusions contained herein are those of the authors and should not be interpreted as necessarily representing the official policies or endorsements, either expressed or implied, of IARPA, DOI/IBC, or the U.S. Government.

\bibliographystyle{IEEEtran}
\bibliography{reference}

\end{document}